\documentclass{article}
\usepackage[final]{neurips_2023}

\usepackage{amsmath}
\usepackage{amssymb}
\usepackage{mathtools}
\usepackage{amsthm}
\usepackage[utf8]{inputenc} 
\usepackage[T1]{fontenc}    
\usepackage[breaklinks]{hyperref}      
\usepackage{url}            
\usepackage{booktabs}       
\usepackage{multirow}       
\usepackage{amsfonts}       
\usepackage{nicefrac}       
\usepackage{microtype}      
\usepackage{xcolor}         
\usepackage{float}

\usepackage{lipsum}

\usepackage{wrapfig}
\usepackage{graphicx}
\usepackage{caption}
\usepackage{multirow,makecell}

\usepackage{xspace}
\usepackage{subcaption}

\usepackage{algorithm}
\usepackage{algorithmic}

\usepackage[capitalize,noabbrev]{cleveref}

\theoremstyle{plain}

\theoremstyle{definition}

\theoremstyle{remark}

\usepackage[textsize=tiny]{todonotes}

\title{Fast Inference of  Mixture-of-Experts Language Models with Offloading}
\author{%
  Artyom Eliseev\\
  Moscow Institute of Physics and Technology\\
  Yandex School of Data Analysis \\
  \texttt{lavawolfiee@gmail.com} \\
  \And
  Denis Mazur \\
  Moscow Institute of Physics and Technology \\
  Yandex \\
  Researchcore \\
  \texttt{denismazur8@gmail.com} \\
}

\begin{document}
\maketitle

\begin{abstract}
With the widespread adoption of Large Language Models (LLMs), many deep learning practitioners are looking for strategies of running these models more efficiently.
One such strategy is to use sparse Mixture-of-Experts (MoE) --- a type of model architectures where only a fraction of model layers are active for any given input.
This property allows MoE-based language models to generate tokens faster than their ``dense'' counterparts, but it also increases model size due to having multiple ``experts''.
Unfortunately, this makes state-of-the-art MoE language models difficult to run without high-end GPUs.
In this work, we study the problem of running large MoE language models on consumer hardware with limited accelerator memory.
We build upon parameter offloading algorithms and propose a novel strategy that accelerates offloading by taking advantage of innate properties of MoE LLMs.
Using this strategy, we build can run Mixtral-8x7B with mixed quantization on desktop hardware and free-tier Google Colab instances.

\end{abstract}
\section{Introduction}
\label{sect:intro}

Many recent advances in natural language processing rely on large pre-trained language models, such as GPT-3 and 4~\cite{brown2020language,openai2023gpt}, Palm \& Gemini~\cite{chowdhery2022palm,gemini} and many others. However, the rapid scientific progress in this area would be impossible without \emph{open-access} LLMs such as LLaMA 1 and 2~\citep{touvron2023llama}, Falcon~\citep{falcon2023}, BLOOM~\citep{scao2022bloom}, OPT~\citep{zhang2022opt}, or NeoX/Pythia~\citep{biderman2023pythia}.
The key advantage of open-access LLMs is that researchers can deploy them \textit{locally} and modify them in ways that would be impossible with proprietary APIs.

Even though LLM parameters are openly available, it is still difficult to use these models due to their sheer size. State-of-the-art open-access language models require multiple high-end GPUs~\footnote{When deployed in 16-bit precision, Falcon-180B needs approximately 360GB, while LLaMA-2 70B requires 140GB of combined accelerator memory.} even for basic inference workloads. To use these LLMs on more affordable hardware setups, one must either compress model parameters~\citep{dettmers2022llm,frantar2022gptq} or offload parameters to a cheaper storage, be it RAM or SSD~\citep{l2l,flexgen}. 

Several recent works modify transformer architecture by introducing sparse Mixture-of-Experts blocks~\citep{moe_first,shazeer2017outrageously}. MoE blocks contain multiple ``experts'' (layers), as well as a ``gating function'' that selects which experts are used on a given input. As a result, the MoE block uses a small portion of all ``experts'' for any single forward pass, allowing for more compute-efficient training~\cite{fedus2021switch, du2022glam}.
Notably, MoEs are among the largest~\cite{fedus2021switch} and among the best~\cite{mixtral} of available LLMs.
While Mixture-of-Experts models can be more efficient than their dense counterparts, many techniques for efficient LLM inference were not designed with MoE in mind and perform suboptimally on modern large language models that use  mixture-of-experts layers.

In this work, we systematically develop techniques for running large MoE language models with limited GPU memory. Our main objective is inferencing (generating tokens) with Mixtral-8x7B-Instruct --- a MoE-based chat assistant --- on a desktop-grade hardware where only a fraction of experts fit into the accelerator memory. To that end:\begin{itemize}
    \item we observe how MoE language model accesses its experts between tokens, and find several regularities: i) some experts are reused between adjacent tokens and ii) the model hidden states of early layers already ``know'' which experts are to be used at subsequent layers.
    
    \item we design a MoE-specific offloading strategy that takes advantage of these regularities: i) it uses LRU cache to significantly reduces GPU-RAM communication, leading to faster generation and ii) it guesses which experts are needed ahead of time to better overlap expert loading with computation.

    \item we consider the specific scenario of running Mixtral-8x7B-Instruct on a T4, RTX 3060 and RTX 3080 Mobile and develop a practical combination of mixed quantization and the proposed offloading algorithm to run this model interactively at 2-3 tokens per second depending on the hardware. The source code with our implementation is available online\footnote{\url{https://github.com/dvmazur/mixtral-offloading}}
\end{itemize}

\section{Background \& Related Work}
\subsection{Mixture-of-Experts}\label{sect:related_moe}

The recent surge in MoE language models builds on a relatively old idea~\citep{moe_first, jordan1994hierarchical} of training ensembles of specialized models (``experts'') and a gating function to select the right expert for the task. To achieve specialization, Mixture-of-Experts learn by simultaneously i) training the gating function to choose the best experts and ii) training the experts themselves on samples assigned to them by the gating function. Since then, many different MoE variants emerged, including mixture of SVM models~\citep{moe_svm}, Dirichlet processes~\citep{moe_dirichlet} and various neural networks.

\cite{shazeer2017outrageously} builds on this idea to train a \textit{sparsely gated} Mixture-of-Experts to serve as a language model. The full model consists of a recurrent neural network backbone and a MoE module with up to 131072 experts. When processing a given token, a linear gating function select 4 most suitable experts based on the latest hidden state. The resulting model (including the gating function and experts) is trained end-to-end to minimize cross-entropy, with an additional regularizer to promote equal expert utilization. \cite{shazeer2017outrageously} observed that the MoE model not only improves perplexity, but also learns interpretable expert specializations: some experts would ``specialize'' on prepositions, while others learn to express a particular concept (e.g. speed).

Since then, several lines of work explore Mixture-of-Experts with Transformer-based language models for machine translation~\cite{lepikhin2020gshard}, masked language modeling~\cite{fedus2021switch}, general-purpose LLMs~\cite{du2022glam} and others. Most of these models follow traditional (dense) Transformer architecture for embeddings and attention layers, and only use Mixture for the feedforward (MLP) blocks and use a linear token-level gating function.\nocite{pkm,ryabinin2020crowdsources,lewis2021base} A common observation across most of these works is that MoE models are cheaper to train and inference~\cite{fedus2021switch,lepikhin2020gshard}, but require more parameters than a dense model with equivalent perplexity.

Pre-trained Mixture-of-Experts LLMs have been openly available for over a year\footnote{\url{https://huggingface.co/google/switch-c-2048}, released in November 15th, 2022}. However, these models seem to have gained less traction than equivalent dense models, arguable because their sheer model size (over a trillion parameters) makes them difficult to use. Most recently, Mistral AI released a family of sparse Mixture of Experts models called Mixtral-8x7B with near state-of-the-art performance~\cite{mixtral}. This model has already inspired several follow-up works and practical applications, but it still requires a high-end GPU accelerator.

\subsection{Post-training Quantization of LLMs}\label{sect:related_quantization}

A natural way to circumvent this is to reduce the model size through quantization~\citep{nagel2020up, gholami2021survey,frantar2022gptq}, sparsification~\cite{frantar-sparsegpt,ma2023llmpruner}, factorization~\cite{hsu2022language}, or a combination thereof. These compression types are not specific to LLMs and are based on much older methods outside the scope of our work\footnote{To learn more about these methods, please refer to surveys such as~\cite{gholami2021survey,pqsurvey}}. However, recent works found that there are unique challenges to quantizing very large transformer-based language models due to emergent outliers\cite{dettmers2022llm,lin2023awq,dettmers2023spqr}.

Generally speaking, the optimal compression rate for most LLMs is 4 bits per parameter~\cite{dettmers2022case}. While there are more extreme algorithms for 3- and even 2-bit compression~\cite{chee2023quip,lin2023awq,dettmers2023spqr}, they are typically inferior to choosing a smaller model and quantizing it to around 4 bits. Most recently, there has been several concurrent works for quantizing Mixture-of-Experts models~\citep{moqe,qmoe}.

\subsection{Inference with Parameter Offloading}\label{sect:offloading}

A recent line of work explores inferencing and training large models with limited accelerator memory by ``offloading'' their parameters to another, cheaper memory, such as system RAM or even SSD~\citep{l2l,zerooffload}. This technique works by loading model parameters just-in-time when they are needed for computation. Since most deep learning models use layers in a fixed order, offloading can pre-dispatch the next layer parameters in the background, ahead of time. 

This technique works particularly well when processing large batches of data, during training~\cite{l2l,zerooffload} or large-batch non-interactive inference~\cite{deepspeed_inference,flexgen}, where each layer processes a lot of tokens each time the layer is loaded from RAM. In turn, when doing interactive inference (e.g. as a chat assistants), offloading works significantly slower than on-device inference. This is because interactive inference generates tokens autoregressively, from left to right. This way, the inference system processes one or few tokens at a time, and therefore spends most of the time waiting for next layer's parameters to be loaded.

\subsection{Hardware Setup}\label{sect:hardware}
While our analysis is not specific to any hardware setup, we target the hardware specifications of cheap / free-tier cloud instances~\cite{colab} and the upper half of gaming computers~\cite{steamstatsoct}: i) enough system memory to hold model parameters, ii) a GPU with 11-16GB VRAM and iii) host-to-device communication at 8-16GB/s (PCIe Gen.3). If we examine popular open-access MoE models (Mixtral-8x7B and switch-c-2048), we find that all non-experts can fit a fraction of available GPU memory.
In turn, the experts that constitute vast majority of model parameters do not fit even with quantization. Finally, even if we could fit the model parameters in memory, running generative inference requires additional memory for layer activations and past attention keys \& values.

\section{Method}\label{sect:method}

In this work, we aim to systematically find the optimal way to inference modern Mixture-of-Experts LLMs on desktop or low-end cloud instances. More specifically, we focus on the task of generating tokens interactively, i.e. generate multiple tokens per second at batch size 1\footnote{As opposed to running a processing a large batch of texts over many seconds, as in ~\cite{flexgen}}.

The generative inference workload consists of two phases: 1) encoding the input prompt and 2) generating tokens conditioned on that prompt. 
The key difference between these two phases is that prompt tokens are encoded in parallel (layer-by-layer), whereas the generation runs sequentially (token-by-token and layer-by-layer). In general, phase 1 works relatively well with existing Mixture-of-Experts algorithms, since each layer can only be loaded once for the entire prompt. In turn, when generating tokens, one must load layer once per each token generated. In practice, this means that inference speed is limited by how fast one can fetch parameters from system memory.

Below, we look for patterns in how the  MoE model loads its experts and propose ways to exploit these patterns to speed up inference time.

\begin{figure}[t]
    \centering
    \vspace{-10px}\includegraphics[width=\textwidth]{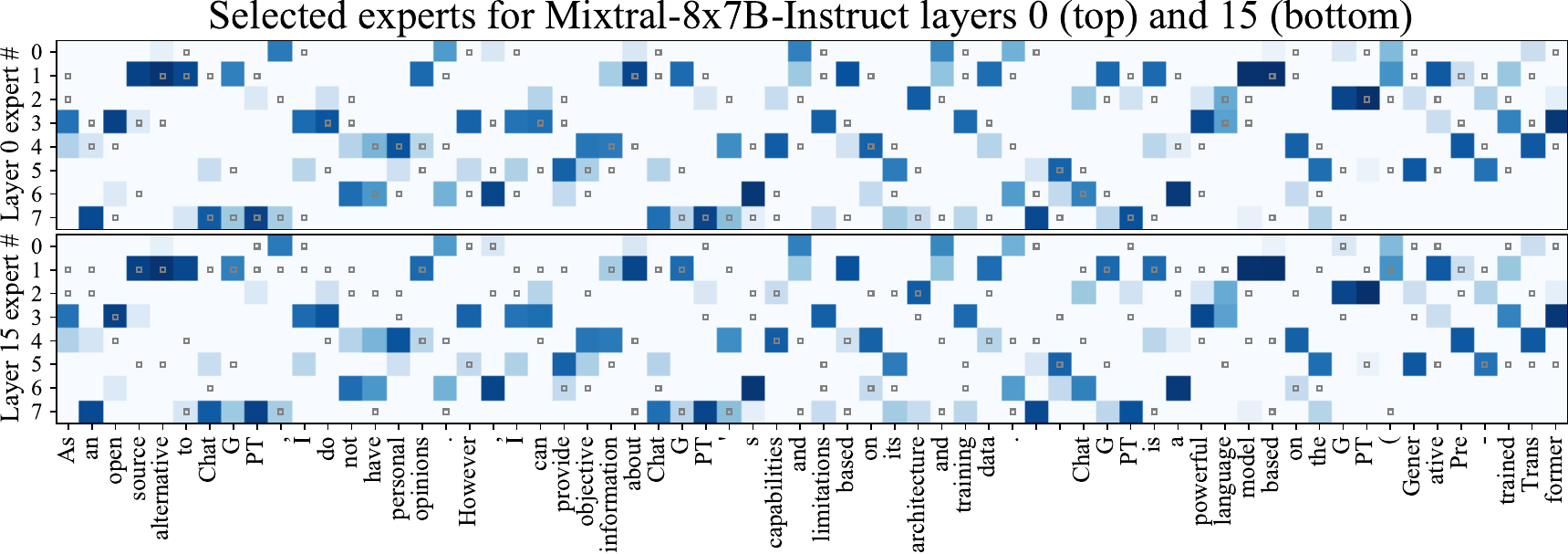}
    \caption{An example of expert loading pattern in Mixtral-8x7B-Instruct for select layers. Blue cells indicate that a certain expert was active when encoding a certain token; deeper blue indicates higher gating weight. Small gray squares show which experts are cached with an LRU cache for $k{=}2$.}\vspace{-10px}
    \label{fig:cache_hits_example}
\end{figure}

\subsection{Expert Locality and LRU caching}\label{sect:method_lru}

As we discussed earlier in Section~\ref{sect:related_moe}, Mixture-of-Experts language models were often observed to assign individual experts to distinct sub-tasks. However, this does not mean that the model uses the same expert over long stretches of tokens. Instead, some experts are active in short sequences of 2-4 tokens, while others are often used with ``gaps'', as shown in Figure~\ref{fig:cache_hits_example}.

To take advantage of this pattern, we can keep active experts in GPU memory as a ``cache'' for future tokens. If the same experts are activated again in future, they will be available instantaneously. Naturally, the number of experts that can be stored this way if very limited by the available GPU memory. For simplicity, we choose to always keep $k$ least recently used experts as a type of LRU cache. If $k$ is greater than the number of active experts, the cache will save experts from multiple previous tokens. For simplicity, we keep the same number of cached experts for each MoE layer.

We illustrate an example of how LRU cache saves experts in Figure~\ref{fig:cache_hits_example} (see caption). LRU is a very simple strategy that does not consider factors like expert activation frequencies, varying cache size between MoE layers, or any sequential patterns in expert activation. However, we found that even this simple strategy can significantly speed up inference for modern Mixture-of-Experts models such as Mixtral-8x7B (see Section~\ref{sect:experiments} for detailed evaluation).

\subsection{Speculative Expert Loading}\label{sect:method_prefetch}

While LRU caching can reduce the average expert loading time, most of the inference time is still spent waiting for the next expert to be loaded. The reason behind this is that, unlike with dense models, MoE offloading cannot effectively overlap expert loading with computation. To understand this problem, let us zoom into the process of generating a single token, layer-by-layer. The full compute workload starts by embedding the previous token via look-up, then alternates between running self-attention and MLP for each transformer block in the model. Finally, the outputs from the last transformer block are used to predict next token logits with a linear projection.

For regular (dense) models, this architecture allows for efficient offloading schedule that pre-loads the next transformer layer ahead of time, while the previous layer is still running. Unfortunately, this schedule is no longer possible for Mixture-of-Experts models, where MoE MLP layers choose which experts to load just-in-time for computation. This is because the system cannot pre-fetch the next layer until it learns which experts should be loaded. Modern open-access MoE language models choose active experts using the final outputs of the previous layer, which means they cannot be pre-fetched them in parallel with previous layer.

While it is not possible\footnote{More specifically, not possible without changing the model architecture, which would require re-training} to pre-reliably prefetch the next set of experts ahead of time, the system could still try to guess the \textit{likely} next experts and load them speculatively, while processing the previous layer. It the guess is correct, it will speed up the next layer inference; if not, it can load the actual next layer's experts later. In other words, this type of speculative loading does not change the final model predictions, but may reduce latency if the guess is accurate enough.

While analyzing modern MoE models, we found that it is possible to get an accurate guess of next layer's experts by \textbf{applying next layer's gating function to previous layer's hidden states} --- or, more specifically, to the same hidden states that are used by previous MoE layer's gating function. This heuristic relies on the fact that transformer layers are residual, i.e. each layer adds to the previous hidden states instead of re-computing them from scratch. This architecture introduces an inductive bias such that any layer's hidden states into a decent estimate of next layer's hidden states.

\subsection{System Design \& Implementation Details}\label{sect:method_system}

In this section, we describe practical design considerations and implementation details that we used for inferencing MoE language models on consumer and low-end cloud hardware. Our system design combines the caching \& prefetching techniques and a mixed MoE quantization scheme .

\textbf{MoE quantization.} As we described earlier in Section~\ref{sect:related_quantization}, there are multiple weight quantization algorithms optimized for LLMs. Model compression has a natural synergy with offloading because compressed models take less time to load onto GPU.
In our experitments, we also observed that MoE models get better quality-size trade-offs when \textbf{quantizing experts to a lower bitwidth}, while keeping all non-expert layers at 4-bit.

We use Half Quadratic Quantization (HQQ)~\citep{badri2023hqq} --- a data-free quantization algorithm that supports a variety of bit rates.  However, we chose this algorithm only for convenience, because it was already well tested for Mixtral models. Since our analysis does not rely on any specific choice of quantization, we believe that if we chose another quantization algorithm (e.g. GPTQ or AWQ) our conclusions would be similar. In our early experiments, we also tried the sub-1-bit quantization from QMoE~\cite{qmoe} that worked well on the Switch-c-2048 model. However, we found that sub-1-bit compression caused too significant a loss in perplexity for Mixtral-8x7B models.

\textbf{Expert Offloading.} As described earlier, we use LRU cache with an equal number $k$ of cached experts per layer. For Mixtral-8x7B, we use $k{=}2$ for 12GB GPUs and $k{=}4$ for 16GB ones. We trigger speculative expert loading immediately after the system finished loading all experts for the current layer. The speculative expert loading fetches $1-2$ most likely experts. The newly loaded experts do not replace the currently cached experts. If a speculatively loaded expert was later used during next layer inference, it will replace the least recently used expert from the next layer's cache.

Many consumer devices and free-tier cloud instances have limited host RAM that cannot fit the entire model\footnote{Notably, Google Colab RAM cannot fit Mixtral-8x7B with a reasonable compression rate}. In these cases, the experts must be split between host and device memory. To support this, our implementation of expert LRU cache splits experts between host and GPU devices. When loading and expert to the GPU cache, the system also offloads the least recently used on-device expert back to RAM so as to preserve memory parity.

To speed up offloading in practice, we allocate all expert parameters in a contiguous memory buffer that can be moved as a single host-to-device copy. For host-side (RAM) experts, we pin\footnote{This corresponds to $\texttt{tensor.pin\_memory()}$ command in PyTorch.} this memory buffer for faster communication. Our implementation additionally allocates $b{=}4$ on-device buffers used to copy and prefetch experts asynchronously, without modifying existing experts. These buffers are shared between all MoE layers to reduce memory footprint.
Overall, the system requires $\text{num\_layers} \times \text{num\_experts}$ expert memory buffers split between host and device memory and $b{=}4$ temporary buffers, the size of each buffer being equal to a single expert.



\section{Experiments}
\label{sect:experiments}

In this section, we verify our earlier hypotheses about MoE behavior and benchmark the inference latency in different conditions. We focus our evaluations on Mixtral-8x7B and Mixtral-8x7B-Instruct models since they represent the current state of the art among open-access MoE models. We organize this section as follows: Section~\ref{sect:exp_caching} measures the effectiveness of expert caching and pre-loading in isolation, Section~\ref{sect:exp_quantization} compares different model compression algorithms and verifies our hypotheses from Section~\ref{sect:method_system}. Finally, Section~\ref{sect:exp_latency} measures the inference latency in several hardware setups.

\subsection{Expert LRU Cache and Speculative Loading}\label{sect:exp_caching}

In this section, we benchmark the effectiveness of the two expert offloading strategies: LRU caching and and speculative loading, as defined in Sections~\ref{sect:method_lru} and~\ref{sect:method_prefetch} respectively.
For this evaluation, we measure ``expert recall'' --- the fraction of times when an expert needed for inference was already available on GPU.

\definecolor{MyGreen}{rgb}{0.1803921568627451,0.7215686274509804,0.08627450980392157}

\begin{table}[b]
\hspace{-10px}\setlength{\tabcolsep}{3pt}
\begin{tabular}{lccccc}
\multicolumn{6}{c}{}\\
 \toprule
 \bf{\begin{tabular}[c]{@{}c@{}}Attn\\ quant\end{tabular}} & \bf{\begin{tabular}[c]{@{}c@{}}Experts\\ quant\end{tabular}} & \bf{\begin{tabular}[c]{@{}c@{}}Model\\ size, GB\end{tabular}} & \bf{Wiki2} & \bf{C4} & \bf{MMLU} \\
 \midrule

 \multirow{4}{*}{FP16} & FP16 & 86.99 & 3.59 & 6.52 & 70.51\% \\
  & 4-bit & 25.82 & 3.67 & 6.58 & 70.3\% \\
 & 3-bit & 23.21 & 3.96 & 6.78 & 69.32\% \\
 & 2-bit & 19.33 & 4.52 & 7.31 & 66.66\% \\
 \midrule
 \multirow{4}{*}{\color{MyGreen} 4-bit} & FP16 & 85.16 & 3.68 & 6.59 & --- \\
  & 4-bit & 23.99 & 3.76 & 6.66 & 69.11\% \\
 & \color{MyGreen} 3-bit & \color{MyGreen} 21.37 & \color{MyGreen} 4.05 & \color{MyGreen} 6.87 & \color{MyGreen} 68.47\% \\
 & \color{MyGreen} 2-bit & \color{MyGreen} 17.54 & \color{MyGreen} 4.61 & \color{MyGreen} 7.42 & \color{MyGreen} 65.58\% \\
 \bottomrule
 \end{tabular}
 \hfill
  \begin{tabular}{lccccc}
\multicolumn{6}{c}{}\\
 \toprule
 \bf{\begin{tabular}[c]{@{}c@{}}Attn\\ quant\end{tabular}} & \bf{\begin{tabular}[c]{@{}c@{}}Experts\\ quant\end{tabular}} & \bf{\begin{tabular}[c]{@{}c@{}}Model\\ size, GB\end{tabular}} & \bf{Wiki2} & \bf{C4} & \bf{MMLU} \\
 \midrule

 \multirow{4}{*}{3-bit} & FP16 & 85.08 & 3.99 & 6.90 & --- \\
  & 4-bit & 23.92 & 4.06 & 6.97 & 66.54\% \\
 & 3-bit & 21.31 & 4.34 & 7.21 & 65.79\% \\
 & 2-bit & 17.46 & 4.90 & 7.82 & 61.83\% \\
 \midrule
 \multirow{4}{*}{2-bit} & FP16 & 84.96 & 4.98 & 7.92 & --- \\
  & 4-bit & 23.79 & 5.08 & 8.06 & 59.0\% \\
 & 3-bit & 21.18 & 5.36 & 8.34 & 57.67\% \\
 & 2-bit & 17.30 & 5.97 & 9.11 & 55.26\% \\
 \bottomrule
\end{tabular}
\vspace{5px}\caption{Perplexity and model size evaluation of Mixtral-8x7B with different quantization for shared attention (Attn quant) and experts (Experts quant) layers. For comprarison, a Mistral-7B 4-bit quantized model has Wiki2 perplexity 5.03, C4 perplexity 7.56 and MMLU score 61.3\%. See Section 4.2 for details. Green values correspond to the configurations we chose for full system evaluation.}\label{tab:eval}

\end{table}

For this evaluation, we run Mixtral-8x7B-Instruct model on the OpenAssistant dataset~\citep{openassistant}. We test LRU caching by running the model on recorded conversations and measuring the recall (aka ``hit ratio'' from caching perspective) for different cache sizes $k$. Next, we test speculative loading in isolation by ``guessing'' which experts should be loaded (by applying the next layer's gating function on current layer activations), then measuring how often the actual next experts get loaded this way. A recall of 1.0 corresponds to a situation where both (2) Mixtral active experts were pre-fetched. We test speculative loading in three settings: 1, 2 and 10 layers ahead.

\begin{figure}[t]
    \centering
    \includegraphics[width=0.47\textwidth]{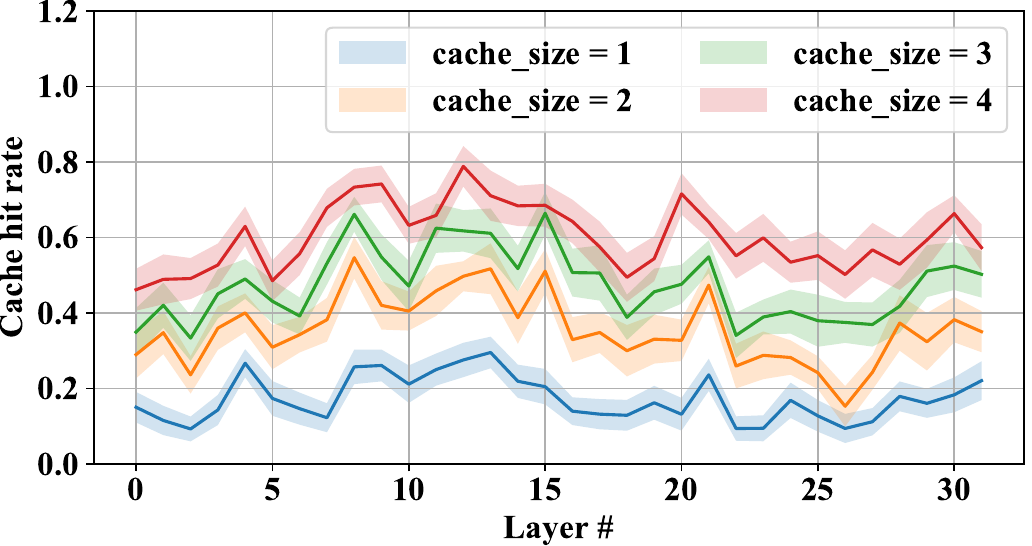}
    \hfill
    \includegraphics[width=0.47\textwidth]{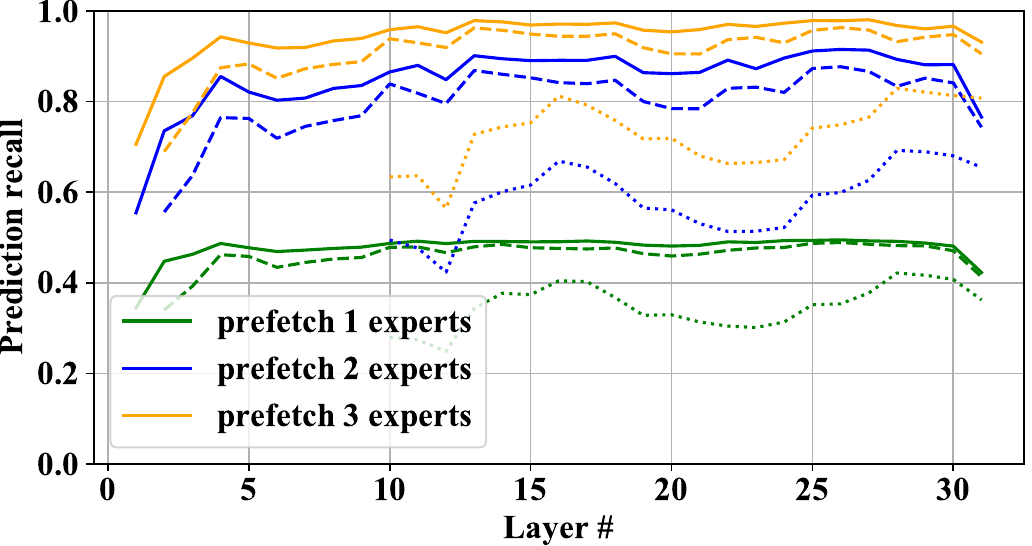}
    \caption{\textbf{(left)} LRU cache hit ratio for different cache size $k$; \textbf{(right)} speculative loading recall when pre-loading a different number of experts. Regular lines represent loading 1 layer ahead; \textbf{dashed} line stands for 2 layers ahead; \textbf{dotted} line is 10 layers ahead.}\vspace{-10px}
    \label{fig:cache_hits}
\end{figure}

\subsection{Mixed MoE Quantization}\label{sect:exp_quantization}

Next, we test how different Quantization schemes affect MoE performance and size. We also use Mixtral-8x7B, but this time, we use non-instruction-tuned variant since it fits better with the available benchmarks. We measure WikiText2 perpliexity~\cite{wikitext103}, C4 perplexity~\cite{C4}, as well as 5-shot MMLU accuracy~\cite{mmlu}. Our objective for this section is to find the best trade off between size and performance for offloading with the target setups. Note that out of 46.7B total parameters in the Mixtral-8x7B model, the experts constitute 45.1B (96.6\%). The rest of the model parameters are allocated to embeddings, self-attention layers, MoE gates and minor layers such as LayerNorm.


As discussed earlier, we use HQQ~\cite{badri2023hqq} data-free quantization algorithm and consider the following quantization schemes:
\begin{enumerate}
    \item FP16 (no quantization)
    \item HQQ 4-bit with group size 64, scale group size 256
    \item HQQ 3-bit with group size 64, scale group size 128
    \item HQQ 2-bit with group size 16, scale group size 128
\end{enumerate}

Note that the actual model size with n-bit quantization is larger than $n$ bits per parameter. This is because the quantized data format also stores quantization scale and zero point for each group of weights. Notably, the above \textbf{2-bit quantization scheme uses, on average, 2.6 bits per parameter} due to a large number of quantization schemes. We also keep embeddings, logits, MoE gates and normalization layers in 16-bit format.

Table~\ref{tab:eval} summarizes our results: overall, it seems advantageous to quantize experts to 3 or 2 bits while keeping attention layers to a higher bitwidth (16 or 4 bits). Based on these evaluations, we chose two quantization schemes (highlighted in green) that offer favourable performance-size trade-offs within the target hardware constraints.


\definecolor{MyGreen}{rgb}{0.1803921568627451,0.7215686274509804,0.08627450980392157}

\begin{table}[t!]
\setlength{\tabcolsep}{3pt}
\hspace{-20px}\begin{tabular}{l|cccc|cccc}
\toprule

\multirow{2}{*}{\bf{Algorithm}} & \multicolumn{4}{c}{\bf{2-bit Experts}} & \multicolumn{4}{|c}{\bf{3-bit Experts}} \\

          & \bf{A100}
          & \bf{3080 Mobile}
          & \bf{3060}
          & \bf{T4 (Colab)}
          & \bf{A100}
          & \bf{3080 Mobile}
          & \bf{3060}
          & \bf{T4 (Cloud)} \\
\midrule
Full algorithm                        & 3.061                     & 2.655                      & 2.278                     & 2.092                           & 2.845                     & 2.475                      & 2.038                     & 1.603                           \\
W/o expert pre-loading            & 2.918                     & 2.227                      & 2.051                     & 1.567                           & 2.683                     & 2.024                      & 1.857                     & 1.365                           \\
W/o LRU cache \& pre-loading   & 2.265                     & 1.758                      & 1.547                     & 1.168                           & 2.055                     & 1.595                      & 1.346                     & 1.061                           \\
Naive offloading (\texttt{accelerate})                  & 1.392                     & 1.059                      & 0.919                     & 0.661                           & 1.246                     & 0.914                      & 1.791                     & 0.580           \\

 \bottomrule

\end{tabular}
\vspace{5px}\caption{Inference speed for Mixtral-8x7B in low-tier , measured in tokens per second.}\label{tab:inferspeed}
\end{table}


\subsection{Practical offloading performance}\label{sect:exp_latency} 

Finally we evaluate the performance of the Mixtral8x7B-Instruct model using the offloading techniquesproposed throughout this report. Based on the perplexity evaluations from the previous section, we chose 4-bit HQQ quantization for the shared attention layers and 2- or 3-bit quantization for experts. We evaluate this system by generating tokens via sampling on OpenAssistant~\citep{openassistant} conversations and measuring the average number of tokens generated per second with batch size 1. For this evaluation, we always sample proportionally to the predicted probabilities, i.e. without temperature or nucleus sampling.

We consider four hardware configurations: a free-tier Colab instance with a T4 GPU (16GB VRAM, PCIe Gen.3), a past generation gaming laptop with RTX 3080 Mobile (16GB, PCIe Gen.4), a mid-range gaming desktop with RTX 3060 (12GB, PCIe Gen.3) and a high-end data-center server with A100-80GB-SXM. Note that the A100 server could run the model without offloading. We use offloading on A100 mostly to provide a reference for other setups. Finally, when evaluating 3-bit models, we use a cloud T4 from Microsoft Azure because the free-tier colab instances did not have enough RAM for this specific configuration. We use $k=2$ for RTX 3060 and $k=4$ for all other GPUs.

As shown in Table~\ref{tab:inferspeed}, all evaluated setups can generate 2-4 tokens per second with the full algorithm. Using pre-loading appears to be most beneficial on RTX 3060, possibly due to lower LRU cache size. Cursiously, RTX 3060 (desktop) performs nearly equally with a much higher end 3080 Mobile. We attribute this to the fact that both GPUs are still bottlenecked by host-to-device bandwidth, limited by the PCIe architecture. Finally, all schemes significantly outperform naive offloading that loads the entire MoE layer.

\section{Conclusion and Future Work}

In this work, we explore strategies for accelerating Mixture-of-Experts based language models on consumer hardware with limited GPU memory. We propose a MoE-centric approach to offloading and explore how mixed quantization affects perplexity and performance on language understanding tasks. We evaluate the proposed strategies and show that they produce a significant increase in generation speed compared to na\"ve approaches on consumer-grade hardware, including free-tier Google Colab.

Our method provides a practical solution for inferencing large MoE language models on resource-constricted hardware, enabling broader access to these powerful models for research and development. As future work, we plan to explore further offloading strategies, based on speculative expert prediction.

\section*{Acknowledgements}

Authors would like to acknowledge \texttt{mobicham@} for helpful discussions on Mixtral quantization.



\bibliographystyle{bib}
\bibliography{bibliography}

\clearpage
\appendix


\end{document}